\documentclass[10pt,twocolumn,letterpaper]{article}

\usepackage{cvpr}
\usepackage{times}
\usepackage{epsfig}
\usepackage{graphicx}
\usepackage{amsmath}
\usepackage{amssymb}
\usepackage{makecell}

\usepackage{adjustbox}
\usepackage{booktabs, colortbl}
\usepackage[table]{xcolor}

\usepackage{url}
\usepackage{footnote}
\makesavenoteenv{tabular}
\makesavenoteenv{table}

\usepackage[utf8]{inputenc}
\usepackage{booktabs, caption, makecell}

\usepackage{threeparttable}

\usepackage[pagebackref=true,breaklinks=true,letterpaper=true,colorlinks,bookmarks=false]{hyperref}

\cvprfinalcopy 

\def\cvprPaperID{7357} 

\ifcvprfinal\fi
\begin{document}

\title{Squeeze-and-Attention Networks for Semantic Segmentation}

\author{Zilong Zhong$^{1,4}$, Zhong Qiu Lin$^{2}$, Rene Bidart$^{2}$,  Xiaodan Hu$^{2}$,  Ibrahim Ben Daya$^{2}$, Zhifeng Li$^{5}$,\\
Wei-Shi Zheng$^{1,3,4}$, Jonathan Li$^{2}$, Alexander Wong$^{2}$\\
\small{$^1$School of Data and Computer Science, Sun Yat-Sen Univeristy, China}\\
\small{$^2$University of Waterloo, Waterloo, Canada}\\
\small{$^3$Peng Cheng Laboratory, Shenzhen 518005, China}\\
\small{$^4$Key Laboratory of Machine Intelligence and Advanced Computing, Ministry of Education, China}\\
\small{$^5$Mstar Technologies, Hangzhou, China}\\
\small{\{zlzhong, wszheng\}@ieee.org, \{zq2lin, x226hu, ibendaya, junli, a28wong\}@uwaterloo.ca}}


\maketitle
\thispagestyle{empty}
\begin{abstract}

The recent integration of attention mechanisms into segmentation networks improves their representational capabilities through a great emphasis on more informative features. However, these attention mechanisms ignore an implicit sub-task of semantic segmentation and are constrained by the grid structure of convolution kernels. In this paper, we propose a novel squeeze-and-attention network (SANet) architecture that leverages an effective squeeze-and-attention (SA) module to account for two distinctive characteristics of segmentation: i) pixel-group attention, and ii) pixel-wise prediction. Specifically, the proposed SA modules impose pixel-group attention on conventional convolution by introducing an {\lq{attention}\rq } convolutional channel, thus taking into account spatial-channel inter-dependencies in an efficient manner. The final segmentation results are produced by merging outputs from four hierarchical stages of a SANet to integrate multi-scale contexts for obtaining an enhanced pixel-wise prediction.  Empirical experiments on two challenging public datasets validate the effectiveness of the proposed SANets, which achieves 83.2\% mIoU (without COCO pre-training) on PASCAL VOC and a state-of-the-art mIoU of 54.4\% on PASCAL Context.

 
\end{abstract}


\section{Introduction}

Segmentation networks become the key recognition elements for autonomous driving, medical image analysis, robotic navigation and virtual reality. The advances of segmentation methods are mainly driven by improving pixel-wise representation for accurate labeling. However, semantic segmentation is not fully equivalent to pixel-wise prediction. In this paper, we argue that semantic segmentation can be disentangled into two independent dimensions: pixel-wise prediction and pixel grouping. Specifically, pixel-wise prediction addresses the prediction of each pixel, while pixel grouping emphasizes the connection between pixels. Previous segmentation works mainly focus on improving segmentation performance from the pixel-level but largely ignore the implicit task of pixel grouping \cite{long2015fully, chen2018deeplab, zhao2017pyramid, zhang2018context, boykov2004experimental, boykov2006graph}. 

\begin{figure}[t]
\begin{center}
\includegraphics[width=1.0\linewidth]{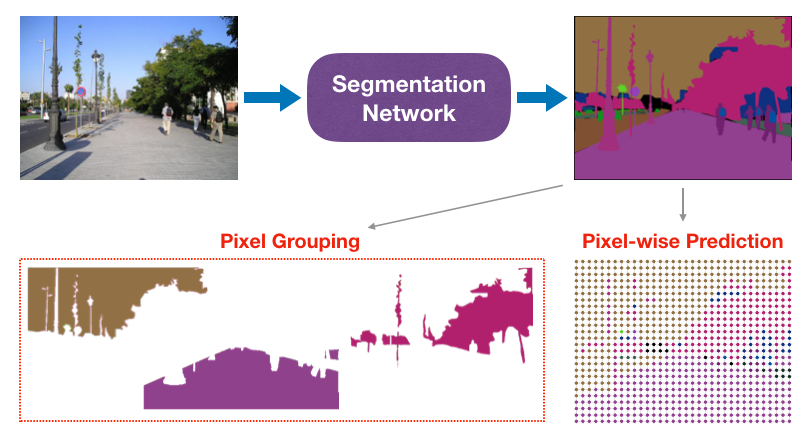}
\end{center}
   \caption{Semantic segmentation can be disentangled into two sub-tasks: explicit pixel-wise prediction and implicit pixel grouping. These two tasks separate semantic segmentation from image classification. Motivated by designing a module that accounts for pixel grouping, we design a novel squeeze-and-attention (SA) module along with a SANet to improve the performance of dense prediction and account for the largely ignored pixel grouping.}
\label{fig1}
\end{figure}

The largely ignored task of pixel grouping can be discovered by disentangling semantic segmentation into two sub-tasks. As shown in Figure \ref{fig1}, the first sub-task requires precise pixel-wise annotation and introduces spatial constraints to image classification. Recent segmentation models achieved significant advances by aggregating contextual features using pyramid pooling and dilated convolution layers for pixel-wise labeling \cite{zhao2017pyramid, chen2018deeplab}. However, the grid structures of these kernels restrict the shapes of spatial features learned in segmentation networks. The feature aggregation strategy enhances pixel-wise prediction results, but the global perspective of understanding images remains under-exploited. 

To this end, we introduce the second sub-task of pixel grouping that directly encourages pixels that belong to the same class being grouped together without spatial limitation. Pixel grouping involves translating images sampled from a range of electromagnetic spectrum to pixel groups defined in a task-specific semantic spectrum, where each entry of the semantic spectrum corresponds to a class. Motivated by designing a module that accounts for pixel grouping, we design a novel squeeze-and-attention (SA) module to alleviate the local constraints of convolution kernels. The SA module contains down-sampled but not fully squeezed attention channels to efficiently produce non-local spatial attention, while avoiding the usage of heavy dilated convolution in output heads. Specifically, An attention convolution are used to generate attention masks because each convolution kernel sweeps across input feature maps. Different from SE modules \cite{hu2017squeeze} that enhance backbones, SA modules integrate spatial attentions and are head units, the outputs of which are aggregated to improve segmentation performance. The spatial attention mechanism introduced by the SA modules emphasizes the attention of pixel groups that belong to the same classes at different spatial scales. Additionally, the squeezed channel works as global attention masks.

We design SANets with four SA modules to approach the above two tasks of segmentation. The SA modules learn multi-scale spatial features and non-local spectral features and therefore overcome the constraints of convolution layers for segmentation. We use dilated ResNet \cite{he2016deep} and Efficient Nets \cite{tan2019efficientnet} as backbones to take advantage of their strong capacity for image recognition. To aggregate multi-stage non-local features, we adopt SA modules on the multi-stage outputs of backbones, resulting in better object boundaries and scene parsing outcomes. This simple but effective innovation makes it easier to generalize SANets to other related visual recognition tasks. We validate the SANets using two challenging segmentation datasets: PASCAL context and PASCAL VOC 2012 \cite{everingham2015pascal, zhou2016semantic, zhou2017scene}. 

The contributions of this paper are three-fold: 
\begin{itemize}
    \item We disentangle semantic segmentation into two sub-tasks: pixel-wise dense prediction and pixel grouping.
    \item We design a squeeze-and-attention (SA) module that accounts for both the multi-scale dense prediction of individual pixels and the spatial attention of pixel groups. 
    \item We propose a squeeze-and-attention network (SANet) with multi-level heads to exploit the representational boost from SA modules, and to integrate multi-scale contextual features and image-level categorical information.
\end{itemize}

\begin{figure}
\begin{center}
\includegraphics[width=0.9\linewidth]{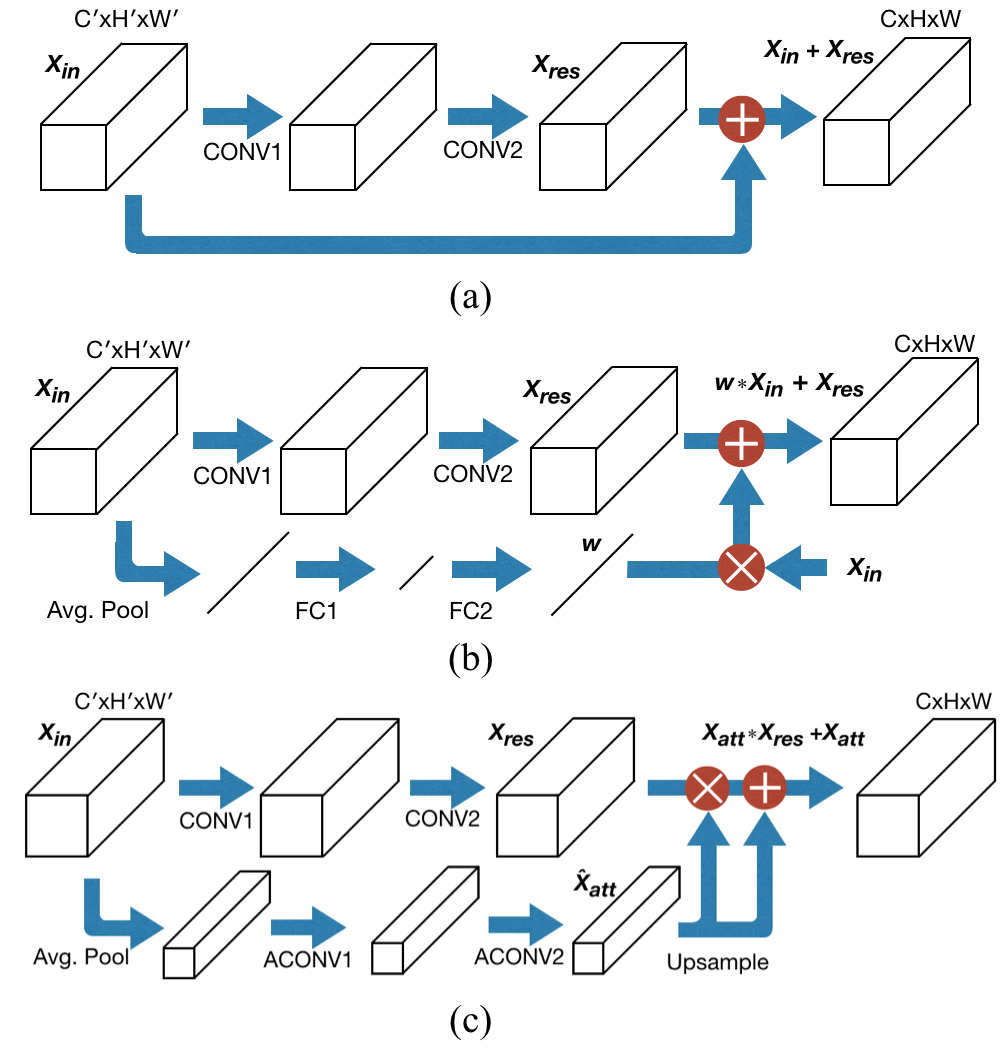}
\end{center}
   \caption{(a) Residual Block; (b) Squeeze-and-excitation (SE) module; (c) Squeeze-and-attention (SA) module; and  For simplicity, we show convolution (CONV), fully connected (FC), average pooling (Avg. Pool) layers, while omitting normalization and activation layers. The SA module has a similar structure as the SE module that contains an additional path to learn weights for re-calibrating channels of output feature maps $X_{out}$.  The difference lies in that the attention  channel of SA modules uses average pooling to down sample feature maps but not fully squeeze as in the SE modules. Therefore, we term this channel the attention convolution (ACONV) channel.}
\label{fig2}
\end{figure}


\section{Related Works}

\textbf{Multi-scale contexts.} Recent improvements for semantic segmentation have mostly been made possible by incorporating multi-scale contextual features to facilitate segmentation models to extract discriminative features. a Laplacian pyramid structure is introduced to combine multi-scale features\cite{ghiasi2016laplacian} introduced. 
A multi-path RefineNet explicitly integrate features extracted from multi-scale inputs to boost segmentation outputs. Encoder-decoder architectures have been used to fuse features that have different levels of semantic meaning \cite{badrinarayanan2015segnet, noh2015learning}. The most popular methods adopt pooling operations to collect spatial information from different scales \cite{zhao2017pyramid, chen2018deeplab}. Similarly, EncNet employs an encoding module that projects different contexts in a Gaussian kernel space to encode multi-scale contextual features  \cite{zhang2018context}. Graphical models like CRF and MRF are used to impose smoothness constraints to obtain better segmentation results \cite{zheng2015conditional, liu2015parsenet, arnab2016higher}. Recently, a gather-excite module is designed to alleviate the local feature constraints of classic convolution by gathering features from long-range contexts \cite{hu2018gather}. We improve the multi-scale dense prediction by merging outputs from different stages of backbone residual networks.

\textbf{Channel-wise attention.} Selectively weighting the channels of feature maps effectively increases the representational power of conventional residual modules. A good example is the squeeze-and-excitation (SE) module because it emphasizes attention on the selected channels of feature maps. This module significantly improves classification accuracy of residual networks by grouping related classes together \cite{hu2017squeeze}. EncNet also uses the categorical recognition capacity of SE modules \cite{zhang2018context}. Discriminative Feature Network (DFN) utilize the channel-weighting paradigm in its smooth sub-network.  \cite{lin2018multi}.

Although re-calibrating the spectral weights of feature map channels has been proved effective for improving the representational power of convolution layers, but the implementation (e.g. squeeze-and-excitation modules) leads to excessive model parameters. In contrast to SE module \cite{hu2017squeeze}, we design a novel squeeze-and-attention (SA) module with a down-sampled but not fully squeezed convolutional channel to produce a flexible module. Specifically, this additional channel generates categorical specific soft attention masks for pixel grouping, while adding scaled spatial features on top of the classical convolution channels for pixel-level prediction.

\textbf{Pixel-group attention.} The success of attention mechanism in neural language processing foster its adoption for semantic segmentation. Spatial Transform Networks explicitly learn spatial attention in the form of affine transformation to increase feature invariance \cite{jaderberg2015spatial}. Since machine translation and image translation share many similarities, RNN and LSTM have been used for semantic segmentation by connecting semantic labeling to translation \cite{zheng2015conditional, lin2018multi}.
\cite{chen2016attention} employed a scale-sensitive attention strategy to enable networks to focus on objects of different scales. \cite{zhao2018psanet} designed a specific spatial attention propagation mechanism, including a collection channel and a diffusion channel. \cite{wang2018non} used self-attention masks by computing correlation metrics. \cite{hu2018gather} designed a gather-and-excite operation via collecting local features to generate hard masks for image classification. Also, \cite{wang2018multi} has proved that not-fully-squeezed module is effective for image classification with marginal computation cost. Since the weights generated by spatially-asymmetric recalibration (SAR) modules are vectors, they cannot be directly used for segmentation.Different from exiting attention modules, we use the down-sampled channels that implemented by pooling layers to aggregate multi-scale features and generate soft global attention masks simultaneously. Therefore, the SA models enhance the objective of pixel-level dense prediction and consider the pixel-group attention that has largely been ignored.



\begin{figure*}
\begin{center}
\includegraphics[width=0.9\linewidth]{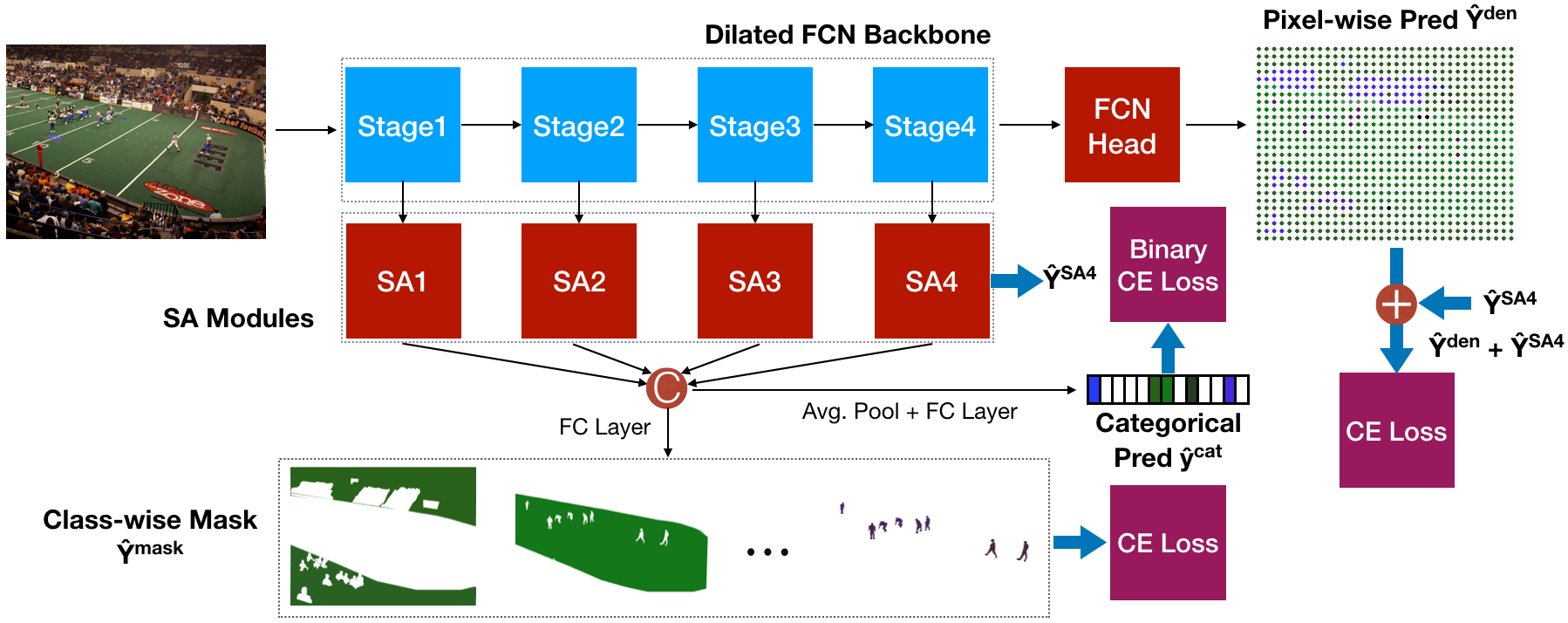}
\end{center}
   \caption{Squeeze-and-attention Network. The SANet aggregates outputs from multiple hierarchical SA heads to generate multi-scale class-wise masks accounting for the largely ignored pixel grouping task of semantic segmentation. The training of these masks are supervised by corresponding categorical regions in ground truth annotation. Also, the masks are used to guide the pixel-wise prediction, which is the output from a FCN head. In this way, we utilize the pixel-group attention extraction capacity of SA modules and integrate multi-scale contextual features simultaneously.}
\label{fig3}
\end{figure*}

\section{Framework}

Classical convolution mainly focuses on spatial local feature encoding and Squeeze-and-Excitation (SE) modules enhance it by selectively re-weighting feature map channels through the use of global image information\cite{hu2017squeeze}.
Inspired by this simple but effective SE module for image-level categorization, we design a Squeeze-and-Attention (SA) module that incorporates the advantages of fully convolutional layers for dense pixel-wise prediction and additionally adds an alternative, more local form of feature map re-weighting, which we call pixel-group attention. Similar to the SE module that boosts classification performance, the SA module is designed specifically for improving segmentation results.

\subsection{Squeeze-and-excitation module}

Residual networks (ResNets) are widely used as the backbones of segmentation networks because of their strong performance on image recognition, and it has been shown that ResNets pre-trained on the large image dataset ImageNet transfer well to other vision tasks, including semantic segmentation \cite{zhao2017pyramid, chen2018deeplab}. Since classical convolution can be regarded as a spatial attention mechanism, we start from the residual blocks that perform as the fundamental components of ResNets. As shown in Figure \ref{fig2} (a), conventional residual blocks can be formulated as:

\begin{equation}
\boldsymbol{X}_{out} = \boldsymbol{X}_{in} + \boldsymbol{X}_{res} = \boldsymbol{X}_{in} + F(\boldsymbol{X}_{in}; \Theta, \Omega)
\end{equation}
where $F(\cdot)$ represents the residual function, which is parameterized by $\Theta$ and $\Omega$ denotes the structure of two convolutional layers. $X_{in} \in \mathbb{R}^{C'\times H'\times W'} $ and $X_{out} \in \mathbb{R}^{C\times H\times W}$ are input and output feature maps. The SE module improve residual block by re-calibrating feature map channels, It is worth noting that we adopt the updated version of SE module, which perform equivalently to original one in \cite{hu2017squeeze}. As shown in Figure \ref{fig2} (b), the SE module can be formulated as:

\begin{equation}
\boldsymbol{X}_{out} = w * \boldsymbol{X}_{in} + F(\boldsymbol{X}_{in}; \Theta, \Omega)
\end{equation}
where the learned weights $w$ for re-calibrating the channels of input feature map ${X}_{in}$ is calculated as:

\begin{equation}
w = \Phi(W_2 * \sigma(W_1 *  APool(\boldsymbol{X}_{in}))),
\end{equation}
where the $\Phi(\cdot)$ represents the sigmoid function and $\sigma(\cdot)$ denotes the ReLU activation function. First, an average pooling layer is used to {\lq{squeeze}\rq } input feature map $X_{in}$. Then, two fully connected layers parameterized by $W_1$ and $W_2$ are adopted to get the {\lq{excitation}\rq } weights. By adding such a simple re-weighting mechanism, the SE module effectively increases the representational capacity of residual blocks. 

\subsection{Squeeze-and-attention module}

Useful representation for semantic segmentation appears at both global and local levels of an image. At the pixel level, convolution layers generate feature maps conditional on local information, as convolution is computed locally around each pixel. Pixel level convolution lays the foundation of all semantic segmentation modules, and increased receptive field of convolution layers in various ways boost segmentation performance \cite{zhao2017pyramid, zhang2018context}, showing larger context is useful for semantic segmentation.  

At the global image level, context can be exploited to determine which parts of feature maps are activated, because the contextual features indicate which classes likely to appear together in the image. Also, \cite{zhang2018context} shows that the global context provides a broader field of view which is beneficial for semantic segmentation. Global context features encode these areas holistically, rather than learning a re-weighting independently for each portion of the image. However, there remains little investigation into encoding context at a more fine-grained scale, which is needed because different sections of the same image could contain totally different environments.

To this end, we design a squeeze-and-attention (SA) module to learn more representative features for the task of semantic segmentation through a re-weighting mechanism that accounts for both local and global aspects. The SA module expands the re-weighting channel of SE module, as shown in Figure \ref{fig2} (b),  with spatial information not fully squeezed to adapt the SE modules for scene parsing. Therefore, as shown in Figure \ref{fig2} (c), a simple squeeze-attention module is proposed and can be formulated as:

\begin{equation}
\begin{split}
\boldsymbol{X}_{out} 
&= \boldsymbol{X}_{attn} * \boldsymbol{X}_{res} + \boldsymbol{X}_{attn} \\
\end{split}
\end{equation}
where $\boldsymbol{X}_{attn}$ = $Up(\sigma(\boldsymbol{\hat{X}}_{attn})) $ and $Up(\cdot)$ is a up-sampled function to expand the output of the attention channel: 

\begin{equation}
\boldsymbol{\hat{X}}_{attn} = F_{attn}(APool(\boldsymbol{X}_{in}); \Theta_{attn}, \Omega_{attn})
\end{equation}
where $\hat{X}_{attn}$ represents the output of the attention convolution channel $F_{attn}(\cdot)$, which is parameterized by $\Theta_{attn}$ and the structure of attention convolution layers $\Omega_{attn}$. A average pooling layer $APool(\cdot)$ is used to perform the not-fully-squeezed operation and then the output of the attention channel $\hat{X}_{attn}$ is up-sampled to match the output of main convolution channel $X_{res}$.

In this way, the SA modules extend SE modules with preserved spatial information and the up-sampled output of the attention channel ${X}_{attn}$ aggregates non-local extracted features upon the main channel.

\begin{figure}[t]
\begin{center}
\includegraphics[width=1.0\linewidth]{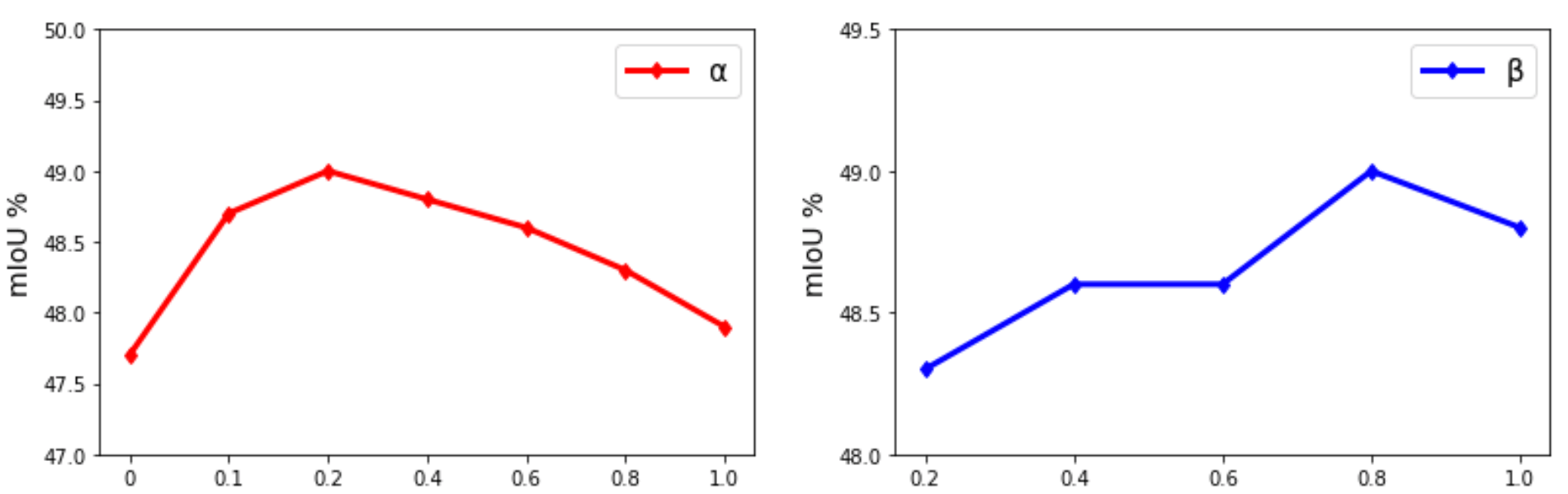}
\end{center}
   \caption{Ablation study of $\alpha$ and $\beta$ that weight the categorical loss and dense prediction loss, respectively. We test SANets using ResNet50 as backbones and train 20 epochs for each case. Left: mIoUs of SANets with fixed $\beta = 0.8$ for selecting $\alpha$. Right mIoUs of SANets with fixed $\alpha = 0.2$ for selecting $\beta$. }
\label{fig4}
\end{figure}



\subsection{Squeeze-and-attention network}

We build a SA network (SANet) for semantic segmentation on top of the SA modules. Specifically, we use SA modules as heads to extract features from the four stages of backbone networks to fully exploit their multi-scale. As illustrated in Figure \ref{fig3}, the total loss involves three parts: dense loss(CE loss), mask loss(CE loss), and categorical loss(binary CE loss). $y_{nj}$ is the average pooled results of $Y^{den}$" Therefore, the total loss of SANets can be represented as: 

\begin{equation}
L_{SANet} = L_{mask} + \alpha * L_{cat} + \beta * L_{den}  
\end{equation}
where $\alpha$ and $\beta$ are weighting parameters of categorical loss and auxiliary loss, respectively. Each component of the total loss can be formulated as follows: 

\begin{equation}
L_{mask} = \dfrac{1}{N \times M}\sum\nolimits_{n=1}^{N} \sum\nolimits_{i=1}^{M}\sum\nolimits_{j=1}^{C} Y_{nij}\log{\hat{Y}_{nij}^{mask}}
\end{equation}
\begin{equation}
\begin{split}
L_{cat} = \dfrac{1}{N}\sum\nolimits_{n=1}^{N} \sum\nolimits_{j=1}^{C}  y_{nj}\log{\hat{y}_{nj}^{cat}} \\
+ (1-y_{nj})\log{(1-\hat{y}_{nj}^{cat})}
\end{split}
\end{equation}
\begin{equation}
L_{den} = \dfrac{1}{N \times M}\sum\nolimits_{n=1}^{N} \sum\nolimits_{i=1}^{M}\sum\nolimits_{j=1}^{C} Y_{nij}\log{\hat{Y}_{nij}^{den}}
\end{equation}
where N is number of training data size for each epoch, M represents the spaital locations, and C denotes the number of classes for a dataset. $\hat{Y}_{nij}$ and $Y_{nij}$ are the predictions of SANets and ground truth, $\hat{y}_{nj}$ and $y_{nj}$ are the categorical predictions and targets to calculate the categorical loss $L_{cat}$. The $L_{cat}$ takes a binary cross entropy form. $L_{mask}$ and $L_{den}$ are typical cross entropy losses. The auxiliary head is similar to the strategy of deep supervision \cite{zhao2017pyramid, zhang2018context}, but its input comes from the fourth stage of backbone ResNet instead of the commonly used third stage. The prediction of SANets integrates the pixel-wise prediction and is regularized by the fourth SA feature map. Hence, the regularized dense segmentation prediction of a SANet is $\hat{Y}^{den} + \hat{Y}^{SA4}$.


\begin{figure}[t]
\begin{center}
\includegraphics[width=1.0\linewidth]{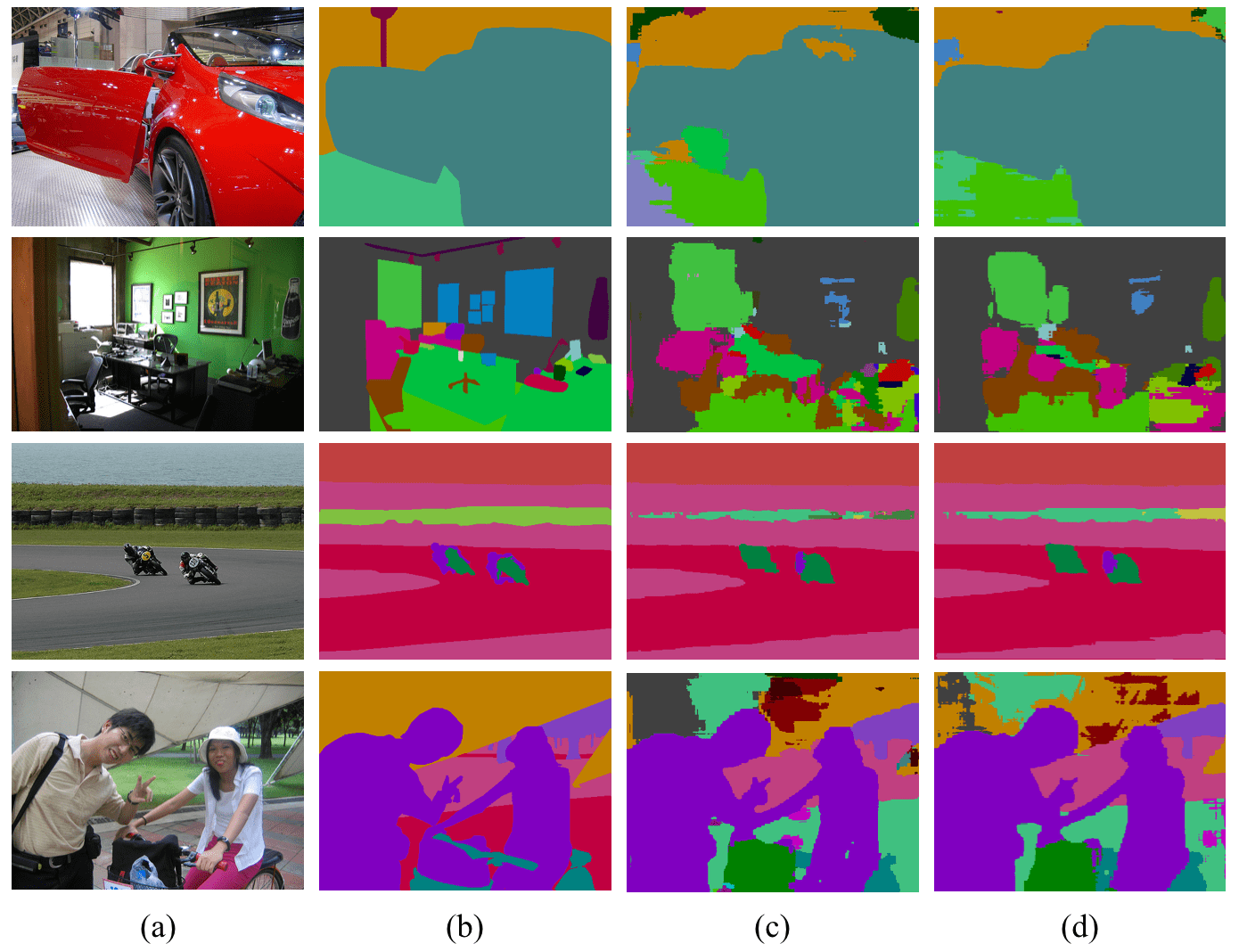}
\end{center}
   \caption{Sample semantic segmentation results on PASCAL Context validation set. Example of semantic segmentation results on PASCAL VOC validation set. (a) Raw images. (b) Groud truth images. (c)  Results of a FCN baseline. (d) Results of a SANet. SANet generates more accurate results, especially for object boundaries. The last raw shows a failed example with relative complex contexts, which bring challenges for segmentation models.}
   
\label{fig5}
\end{figure}


Dilated FCNs have been used as the backbones of SANets. Suppose that the input image has a size of $3 \times 512 \times 512$. The main channel of SA modules has the same channel numbers as their attention counterparts and the same spatial sizes as the input features. Empirically, we reduce the channel sizes of inputs to a fourth in both main and attention channels, set the downsample (max pooling) and upsample ratio of attention channels to 8, and set the channel number of the intermediate fully connected layer of SE modules to 4 in both datasets. We adopt group convolution using 2 groups for the first convolution operations in both main and attention channels. Also, we adapt outputs of SA heads to the class number of segmentation datasets. 

\begin{table}
\begin{center}
\begin{tabular}{c c c c c | c c}
\Xhline{3\arrayrulewidth}
Model & Backbone & SA & Cat & Den & PAcc & mIoU \\
\hline\hline
FCN & Res50 & & & & 74.5 & 43.2 \\
SANet & Res50 & \checkmark & & & 77.2 & 49.2\\ 
SANet & Res50 & \checkmark & \checkmark & & 79.0 & 50.7\\
SANet & Res50 & \checkmark & \checkmark & \checkmark & 79.3 & 51.9\\
SANet & Res101 & \checkmark & \checkmark & \checkmark & 80.6 & 53.0\\
\hline
SANet & EffNet-b7 & \checkmark & \checkmark & \checkmark & 81.6 & 55.3\\
\hline
\end{tabular}
\end{center}
\caption{Ablation study results of SANets on PASCAL Context dataset (59 classes without background). SA: Squeeze-and-attention heads. Cat: Categorical loss. Den: Dense prediction Loss. PAcc: Pixel accuracy (\%). mIoU: Mean intersection of union (\%). }
\label{table2}
\end{table}

\section{Experimental Results}

In this section, we first compare SA module to SE modules, then conduct an ablation study using the PASCAL Context \cite{mottaghi2014role} dataset to test the effectiveness of each component of the total training loss, and further validate SANets on the challenging PASCAL VOC dataset \cite{everingham2010pascal}. Following the convention for scene parsing \cite{chen2018deeplab, zhang2018context}, we paper both mean intersection and union (mIoU) and pixel-wise accuracy (PAcc) on PASCAL Context, and mIoU only on PASCAL VOC dataset to assess the effectiveness of segmentation models.

\subsection{Implementation}

We use Pytorch \cite{paszke2017automatic} to implement SANets and conduct ablation studies. For the training process, we adopt a poly learning rate decreasing schedule as in previous works \cite{zhao2017pyramid, zhang2018context}. The starting learning rates for PASCAL Context and PASCAL VOC are 0.001 and 0.0001, respectively. Stochastic gradient descent and poly learning rate annealing schedule are adopted for both datasets. For PASCAL Context dataset, we train SANets for 80 epochs.
As for the PASCAL VOC dataset, we pretrain models on the COCO dataset. Then, we train networks for 50 epochs on the validation set. We adopt the ResNet50 and ResNet101 as the backbones of SANets because these networks have been widely used for mainstream segmentation benchmarks. We set the batch-size to 16 in all training cases and use sync batch normalization across multiple gpus recentely implemented by \cite{zhang2018context}. We concatenate four SA head outputs to exploit the multi-scale features of different stages of backbones and also to regularize the training of deep networks.

\begin{table}
\begin{center}
\begin{tabular}{c c | c}
\Xhline{3\arrayrulewidth}
Model & Backbone  & mIoU \\
\hline\hline
FCN \cite{long2015fully} &   & 37.8 \\

CRF-RNN\cite{zheng2015conditional} &  & 39.3\\ 
ParseNet\cite{liu2015parsenet} & & 40.4\\
BoxSup\cite{dai2015boxsup} & & 40.5\\
HighOrder-CRF\cite{arnab2016higher} & & 41.3\\
Piecewise\cite{lin2016efficient} & & 43.3\\
Deeplab-v2\cite{chen2018deeplab} & ResNet101 & 45.7\\
RefineNet\cite{lin2017refinenet} & ResNet152  & 47.3\\
EncNet\cite{zhang2018context} & ResNet101  & 51.7\\
\hline
SANet (ours) & ResNet101 & \cellcolor{gray!20}52.1\\
SANet (ours) & EffNet-b7 & \cellcolor{gray!30}\textbf{54.4}\\
\hline
\end{tabular}
\end{center}
\caption{Mean intersection over union (\%) results on PASCAL Context dataset (60 classes with background).}
\label{table3}
\end{table}

\begin{table}
\begin{center}
\begin{tabular}{c c | c}
\Xhline{3\arrayrulewidth}
Model & PAcc  & mIoU \\
\hline\hline
FCN50  & 76.2  & 44.9 \\
FCN101  & 76.7  & 45.6 \\
\hline
FCN50-SE  & 76.0  & 44.6 \\
FCN101-SE  & 76.6  & 45.7 \\
\hline
SANet50 (ours) & 78.9 & 49.0\\
SANet101 (ours) & \textbf{79.2} & \textbf{50.1}\\
\hline
\end{tabular}
\end{center}
\caption{Pixel accuracy (PAcc) and mIoUs of baseline dilated FCNs, dilated FCNs with SE modules (FCN-SE), and SANets using ResNet50 or ResNet101 as backbones on PASCAL Context. SANet significanly output their SE counterparts and baseline models. Each model is trained for 20 epochs}
\label{table4}
\end{table}

\subsection{Results on PASCAL Context}

The Pascal Context dataset contains 59 classes, 4998 training images, and 5105 test images. Since this dataset is relatively small in size, we use it as the benchmark to design module architectures and select hyper-parameters including $\alpha$ and $\beta$. To conduct an ablation study, we explore each component of SA modules that contribute to enhancing the segmentation results of SANets. 

The ablation study includes three parts. First, we test the impacts of the weights $\alpha$ and $\beta$ of the total training loss. As shown in Figure \ref{fig4}, we test $\alpha$ from 0 to 1.0, and find that the SANet with $\alpha=0.2$ works the best. Similarly, we fix $\alpha=0.2$ to find that $\beta=0.8$ yields the best segmentation performance. Second, we study the impacts of categorical loss and dense prediction loss of in equation (7) using selected hyper-parameters. Table \ref{table2} shows that the SANet, which contains the four dual-usage SA modules, using ResNet50 as the backbone improves significantly (a 2.7\% PAcc and 6.0\% mIoU increase) compared to the FCN baseline. Also, the categorical loss and auxiliary loss boost the segmentation performance. 

We compare SANets with state-of-the-art models to validate their effectiveness, as shown in Table \ref{table3}, the SANet using ResNet101 as its backbone achieves 53.0\% mIoU. The mIoU equals to 52.1\% when including the background class this result and outperforms other competitors. 
Also, we use the recently published Efficient Net (EffNet) \cite{tan2019efficientnet} as backbones. Then, the EffNet version SANet achieved state-of-the-art  54.4\% mIoU that sets new records for the PASCAL Context dataset. Figure \ref{fig5} shows the segmentation results of a dilated ResNet50 FCN and a SANet using the same backbone. In the first three rows, SANets generate better object boundaries and higher segmentation accuracy. However, for complex images like the last row, both models fail to generate clean parsing results. In general, the qualitative assessment is in line with quantitative papers. 

We also validate the effectiveness of SA modules by comparing them with SE modules on top of the baseline dilated FCNs, including ResNet50 and ResNet101. Table \ref{table4} shows that the SANets achieve the best accuracy with significant improvement (4.1\% and 4.5\% mIoU increase) in both settings, while FCN-SE models barely improve the segmentation results.


\begin{figure}
\begin{center}
\includegraphics[width=0.9\linewidth]{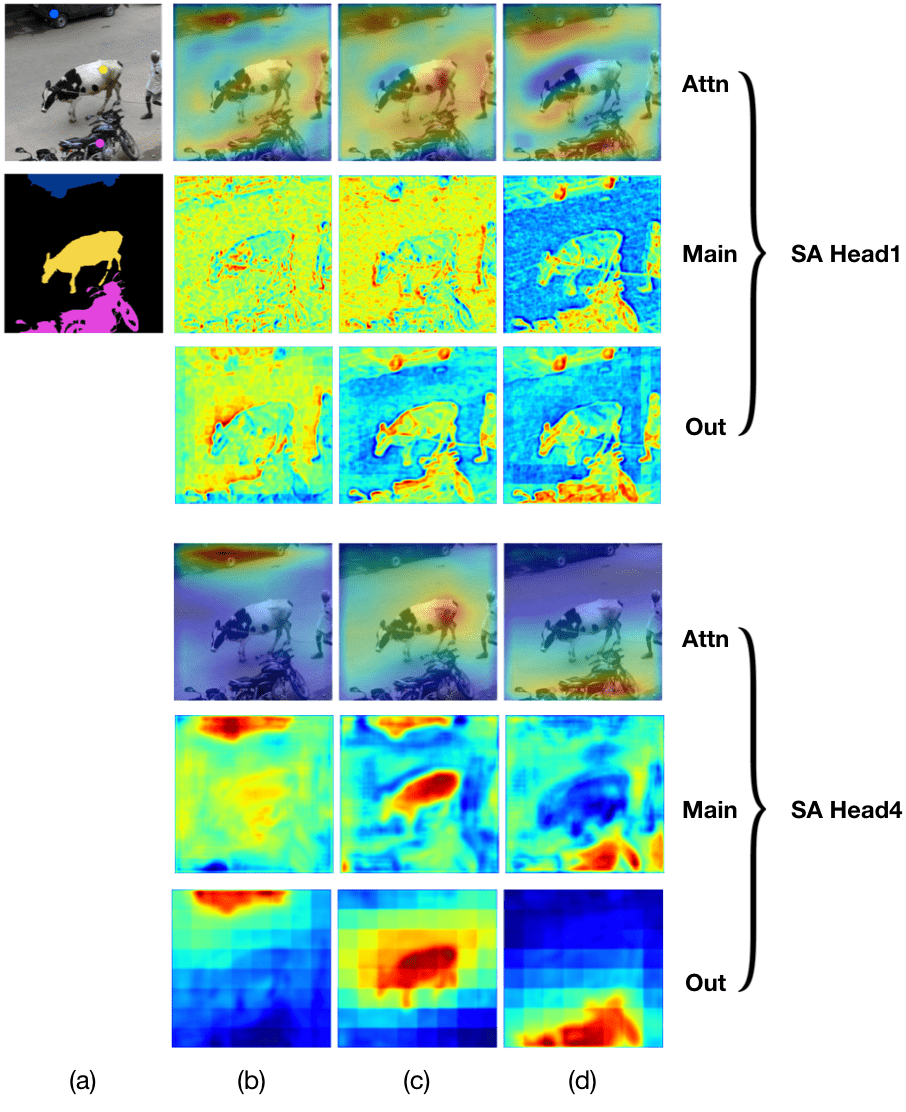}
\end{center}
   \caption{Attention and feature map visualization of SA head1 and head4 of a trained SANet on PASCAL VOC dataset.  For each head, the feature maps of main channel, attention channel, and output are demonstrated. (a) Raw image and its ground truth; the pixel group visualization of (b) blue point; (c) yellow point; and (d) magenta point.}
\label{fig6}
\end{figure}

\subsection{Attention and Feature Maps}

The classic convolution already yields inherent global attention because each convolutional kernel sweeps across spatial locations over input feature maps. Therefore, we visualize the attention and feature maps of a example of PASCAL VOC set and conduct a comparison between Head1 and Head4 within a SANet To better understand the effect of attention channels in SA modules. We use L2 distance to show the attention maps of the attention channel within SA module, and select the most activated feature map channels for the outputs of the main channel within the same SA module. The activated areas (red color) of the output feature maps of SA modules can be regarded as the pixel groups of selected points. For the sake of visualization, we scale all feature maps illustrated in Figure \ref{fig6} to the same size. we select three points (red, blue, and magenta) in this examples to show that the attention channel emphasizes the pixel-group attention, which is complementary to the main channels of SA modules that focus on pixel-level prediction.

Interestingly, as shown in Figure \ref{fig6}, the attention channels in low-level (SA head1) and high-level (SA head4) play different roles. For the low-level stage, the attention maps of the attention channel have broad field of view, and feature maps of the main channel focus on local feature extraction with object boundary being preserved. In contrast, for the high-level stage, the attention maps of the attention channel mainly focus on the areas surrounding selected points, and feature maps of the main channel present more homogeneous with clearer semantic meaning than those of head1.

\begin{table*}
\begin{center}

\begin{threeparttable}
\begin{tabular}{c | c c c c c c c c c c c c | c}
\Xhline{3\arrayrulewidth}
Method   & aero & bike & bird & boat & bottle & bus & car & cat & chair & cow & table & dog & mIoU\\
\hline\hline
FCN  \cite{long2015fully} & 76.8 & 34.2 & 68.9 & 49.4 & 60.3 & 75.3 & 74.7 & 77.6 & 21.4 & 62.5 & 46.8 & 71.8 & 62.2 \\
DeepLabv2 \cite{chen2018deeplab}  & 84.4 & 54.5 & 81.5 & 63.6 & 65.9 & 85.1 & 79.1 & 83.4 & 30.7 & 74.1 & 59.8 & 79.0 & 71.6\\ 
CRF-RNN  \cite{zheng2015conditional}    & 87.5 & 39.0 & 79.7 & 64.2 & 68.3 & 87.6 & 80.0 & 84.4 & 30.4 & 78.2 & 60.4 & 80.5 & 72.0\\
DeconvNet \cite{noh2015learning}   & 89.9 & 39.3 & 79.7 & 63.9 & 68.2 & 87.4 & 81.2 & 86.1 & 28.5 & 77.0 & 62.0 & 79.0 & 72.5\\
GCRF  \cite{vemulapalli2016gaussian}      & 85.2 & 43.9 & 83.3 & 65.2 & 68.3 & 89.0 & 82.7 & 85.3 & 31.1 & 79.5 & 63.3 & 80.5 & 73.2\\
DPN   \cite{liu2015semantic}       & 87.7 & 59.4 & 78.4 & 64.9 & 70.3 & 89.3 & 83.5 & 86.1 & 31.7 & 79.9 & 62.6 & 81.9 & 74.1\\
Piecewise \cite{lin2016efficient}  & 90.6 & 37.6 & 80.0 & 67.8 & 74.4 & 92.0 & 85.2 & 86.2 & 39.1 & 81.2 & 58.9 & 83.8 & 75.3\\
ResNet38 \cite{wu2019wider}    & \cellcolor{gray!20}94.4 & \cellcolor{gray!30}\textbf{72.9} & 94.9 & 68.8 & 78.4 & 90.6 & \cellcolor{gray!20}90.0 & 92.1 & \cellcolor{gray!20}40.1 & 90.4 & 71.7 & 89.9 & 82.5 \\
PSPNet  \cite{zhao2017pyramid}     & 91.8 & \cellcolor{gray!20}71.9 & 94.7 & 71.2 & 75.8 & \cellcolor{gray!20}95.2 &  89.9 & \cellcolor{gray!30}\textbf{95.9} & 39.3 & 90.7 & 71.7 & \cellcolor{gray!20}90.5 & 82.6 \\
DANet \cite{fu2018dual} & -- & -- & -- & -- & -- & -- &  -- & -- & -- & -- & -- & -- & 82.6\\
DFN \cite{yu2018learning} & -- & -- & -- & -- & -- & -- &  -- & -- & -- & -- & -- & -- & 82.7\\
EncNet \cite{zhang2018context} & 94.1 & 69.2 & \cellcolor{gray!30}\textbf{96.3} & \cellcolor{gray!30}\textbf{76.7} & \cellcolor{gray!30}\textbf{86.2} & \cellcolor{gray!30}\textbf{96.3} &  \cellcolor{gray!30}\textbf{90.7} & 94.2 & 38.8 & \cellcolor{gray!20}90.7 & \cellcolor{gray!20}73.3 & 90.0 & \cellcolor{gray!20}82.9\\
\hline
SANet(ours) & \cellcolor{gray!30}\textbf{95.1} & 65.9 & \cellcolor{gray!20}95.4 & \cellcolor{gray!20}72.0 & \cellcolor{gray!20}80.5 & 93.5 &  86.8 & \cellcolor{gray!20}94.5 & \cellcolor{gray!30}\textbf{40.5} & \cellcolor{gray!30}\textbf{93.3} & \cellcolor{gray!30}\textbf{74.6} & \cellcolor{gray!30}\textbf{94.1} &  \cellcolor{gray!30}\textbf{83.2}
 \\
\hline
\end{tabular}

\end{threeparttable}

\end{center}
\caption{Class-wise IoUs and mIoU of PASCAL VOC dataset without pretraining on COCO dataset. The SANet achieves 83.2\% mIoU that outperforms other models and dominates multiple classes. The best two entries of each column are highlighted. To make a fair comparison, modelsuse extra datasets (e.g. JFT) are not included like \cite{chen2017rethinking, luo2017deep, wang2017learning, chen2018encoder}.}
\label{table5}
\end{table*}

\begin{table}
\begin{center}
\begin{threeparttable}
\begin{tabular}{c c | c}
\Xhline{3\arrayrulewidth}
Model & Backbone  & mIoU \\
\hline\hline
CRF-RNN\cite{zheng2015conditional} &  & 74.4\\ 
BoxSup\cite{dai2015boxsup} & & 75.2\\
DilatedNet\cite{yu2015multi} &  & 75.3\\ 
DPN\cite{liu2015semantic} &  & 77.5\\ 
PieceWise\cite{lin2016efficient} & & 78.0\\
Deeplab-v2\cite{chen2018deeplab} & ResNet101 & 79.7\\
RefineNet\cite{lin2017refinenet} & ResNet152   & 84.2\\

PSPNet\cite{zhao2017pyramid} & ResNet101  & 85.4\\

DeeplabV3\cite{chen2018deeplab} &   ResNet101 & 85.7\\
EncNet\cite{zhang2018context} &   ResNet101 & 85.9\\
DFN\cite{yu2018learning} &   ResNet101 & \cellcolor{gray!30}86.2\\
\hline
SANet (ours) & ResNet101 & \cellcolor{gray!20}\textbf{86.1} \\
\hline
\end{tabular}

\end{threeparttable}
\end{center}
\caption{Mean intersection over union (\%) results on PASCAL VOC dataset with pretraining on COCO dataset. The SANet achieves 86.1\% mIoU that is comparable results to state-of-the-art models.  }

\label{table6}
\end{table}

\begin{figure}[t]
\begin{center}
\includegraphics[width=1.0\linewidth]{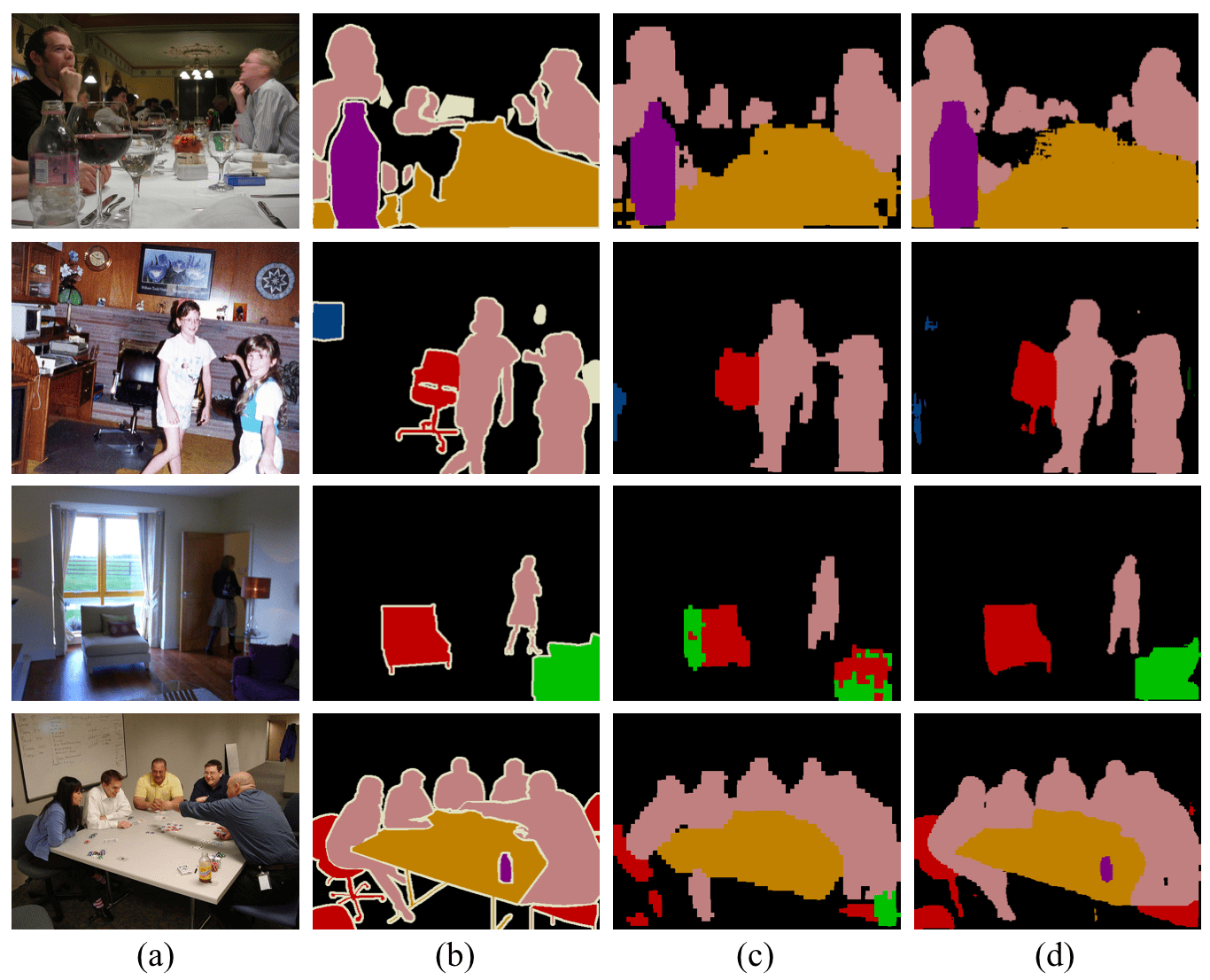}
\end{center}
   \caption{Example of semantic segmentation results on PASCAL VOC validation set. (a) Raw images. (b) Groud truth images. (c) FCN baseline. (d) A SANet. SANet generates more accurate parsing results compared to the baseline.}
\label{fig7}
\end{figure}

\subsection{Results on PASCAL VOC}
The PASCAL VOC dataset \cite{everingham2010pascal} is the most widely studied segmentation benchmark, which contains 20 classes and is composed of 10582 training images, and 1449 validation images, 1456 test images. We train the SANet using augmented data for 80 epochs as previous works \cite{long2015fully, dai2015boxsup}. 

First, we test the SANet without COCO pretraining. As shown in Table \ref{table5}, the SANet achieves 83.2\% mIoU which is higher than its competitors and dominates multiple classes, including aeroplane, chair, cow, table, dog, plant, sheep, and tv monitor. This result validates the effectiveness of the dual-usage SA modules. Models \cite{chollet2017xception, chen2017rethinking} use extra datasets like JFT  \cite{sun2017revisiting} other than PASCAL VOC or COCO are not included in Table \ref{table5}.

Then, we test the the SANet with COCO pretraining. As shown in Table \ref{table6}, the SANet achieves an  evaluated result of 86.1\% mIoU using COCO data for pretraining, which is comparable to top-ranking models including PSPNet \cite{zhao2017pyramid}, and outperforms the RefineNet \cite{lin2017refinenet} that is built on a heavy ResNet152  backbone. Our SA module is more computationally efficient than the encoding module of EncNet \cite{zhang2018context}. As shown in Figure \ref{fig6}, the prediction of SANets yields clearer boundaries and better qualitative results compared to those of the baseline model. 

\subsection{Complexity Analysis}
Instead of pursing SOTA without considering computa-tion costs, our objective is to design lightweight modules for segmentation inspired by this intuition. We use MACs and model parameters to analyze the complexity of SANet. As shown in Table \ref{table7}, both Deeplab V3+ (our implementation) and SAN use ResNet101 backbone and are evaluated on PASCAL VOC dataset to enablea a fair comparison. Without using COCO dataset for pretraining, our SANet surpasses Deeplab V3+ with an increase of 1.7\% mIoU. Compared to heavy-weight models like SDN (238.5M params), SANet achieves slightly under-performed results with less than a fourth number of parameters (55.5M params). The comparison results demonstrate the SANet is effective and efficient.

\begin{table}
\begin{center}
\small\addtolength{\tabcolsep}{-1pt}

\begin{threeparttable}
\begin{tabular}{c| c | c c c}
\Xhline{4\arrayrulewidth}
Model &  Backbone & 
 mIoU & MACs & Params \\
\hline\hline
Dilated FCN & ResNet101 & 78.7 & 162.7G & 42.6M\\
\hline
SDN \cite{fu2019stacked} & DenseNet & 84.2 & -- & 238.5M\\
APCNet \cite{he2019adaptive} & ResNet101 & 83.5 & -- & -- \\ 
Deeplab V3+$^\dagger$ \cite{chen2018encoder} & ResNet101 & 81.5 & 235.6G & 59.5M \\ 
\hline
SANet (ours) & ResNet101 & 83.2 & \textbf{204.7G} & \textbf{55.5M}\\
\hline
\end{tabular}

\begin{tablenotes}\footnotesize
\item[$\dagger$] Our implementation
\vspace{-0.5cm}
\end{tablenotes}

\end{threeparttable}
\end{center}
\caption{MIoUs (\%), Multiply-Accumulate operation per second (MACs) and network parameters (Params) using ResNet101 as backbones evaluated on PASCAL VOC test set without COCO pretraining. We re-implement Deeplab V3+ using dilated ResNet101 as its backbone to enable a fair comparison.}
\label{table7}
\end{table}


\section{Conclusion}

In this paper, we rethink semantic segmentation from two independent dimensions --- pixel-wise prediction and pixel grouping. We design a SA module to account for the implicit sub-task of pixel grouping. The SA module enhances the pixel-wise dense prediction and accounts for the largely ignored pixel-group attention. More importantly, we propose SANets that achieve promising segmentation performance on two challenging benchmarks. We hope that the simple yet effective SA modules and the SANets built on top of SA modules can facilitate the segmentation research of other groups.

\section*{Acknowledgement}
This work was supported partially by the National Key Research and Development Program of China (2018YFB1004903), Research Projects of Zhejiang Lab (No. 2019KD0AB03), International Postdoctoral Exchange Fellowship Program (Talent-Introduction Program), and Google Cloud Platform research credits program. 

{\small
\bibliographystyle{ieee}
\bibliography{bibfile}
}

\end{document}